# Extraction of Line Word Character Segments Directly from Run Length Compressed Printed Text Documents

Mohammed Javed[#1], P. Nagabhushan[#2], B.B. Chaudhuri[*3]

[#]*Department of Studies in Computer Science*

*University of Mysore, Mysore-570006, India*

[1]javedsolutions@gmail.com

[2]pnagabhushan@hotmail.com

[*]*Computer Vision and Pattern Recognition Unit*
*Indian Statistical Institute, Kolkata-700108, India*

[3]bbc@isical.ac.in

***Abstract*- Segmentation of a text-document into lines, words and characters, which is considered to be the crucial preprocessing stage in Optical Character Recognition (OCR) is traditionally carried out on uncompressed documents, although most of the documents in real life are available in compressed form, for the reasons such as transmission and storage efficiency. However, this implies that the compressed image should be decompressed, which indents additional computing resources. This limitation has motivated us to take up research in document image analysis using compressed documents. In this paper, we think in a new way to carry out segmentation at line, word and character level in run-length compressed printed-text-documents. We extract the horizontal projection profile curve from the compressed file and using the local minima points perform line segmentation. However, tracing vertical information which leads to tracking words-characters in a run-length compressed file is not very straight forward. Therefore, we propose a novel technique for carrying out simultaneous word and character segmentation by popping out column runs from each row in an intelligent sequence. The proposed algorithms have been validated with 1101 text-lines, 1409 words and 7582 characters from a data-set of 35 noise and skew free compressed documents of Bengali, Kannada and English Scripts.**

## I. INTRODUCTION

The performance and efficiency of Optical Character Recognition (OCR), Text Document Analysis (TDA), Pattern Recognition and Analysis (PRA) systems depend largely on the accuracy of pre-processing stage of segmentation extracting lines, words and characters respectively. Document segmentation is the process of partitioning document image into physical units of pixels placed in some compactness as defined apriori. There have been several efforts in the literature of document segmentation which are detailed in [1] and [2]. However, these state-of-the-art techniques require that documents should be in uncompressed form. In real life, fax machines [3], Xerox machines, and digital libraries use compressed form to provide better transmission and storage efficiency. To perform any operation, it is required to decompress the document and process further. Thus decompression has been an unavoidable pre-requisite which indents extra computation time and buffer space. Therefore, it would be appropriate to think of performing segmentation operations intelligently straight in their compressed formats.

Handling compressed documents directly for applications in image analysis and pattern recognition, is a challenging goal. The initial idea of working with compressed data was envisioned by a few researchers in early 1980's [4], [5]. Run-Length Encoding (RLE), a simple compression method was first used for coding pictures [6] and television signals [7]. There are several efforts in the direction of directly operating on document images in compressed domain. Operations like image rotation [8], connected component extraction [9], skew detection [10], page layout analysis [11] are reported in the literature related to run-length information processing. There are also some initiatives in finding document similarity [12], equivalence [13] and retrieval [14]. One of the recent work using run-length information is to perform morphological related operations [15]. In most of these works, they use either run-length information from the uncompressed image or do some partial decoding to perform the operations. However to the best of our knowledge, there has been no effort by the document image analysis community to segment a compressed text-documents into lines, words and characters, which is considered to be the first step towards developing an OCR for compressed documents. The overall architecture for developing an OCR or Image Analysis for compressed documents is shown in Fig-1.

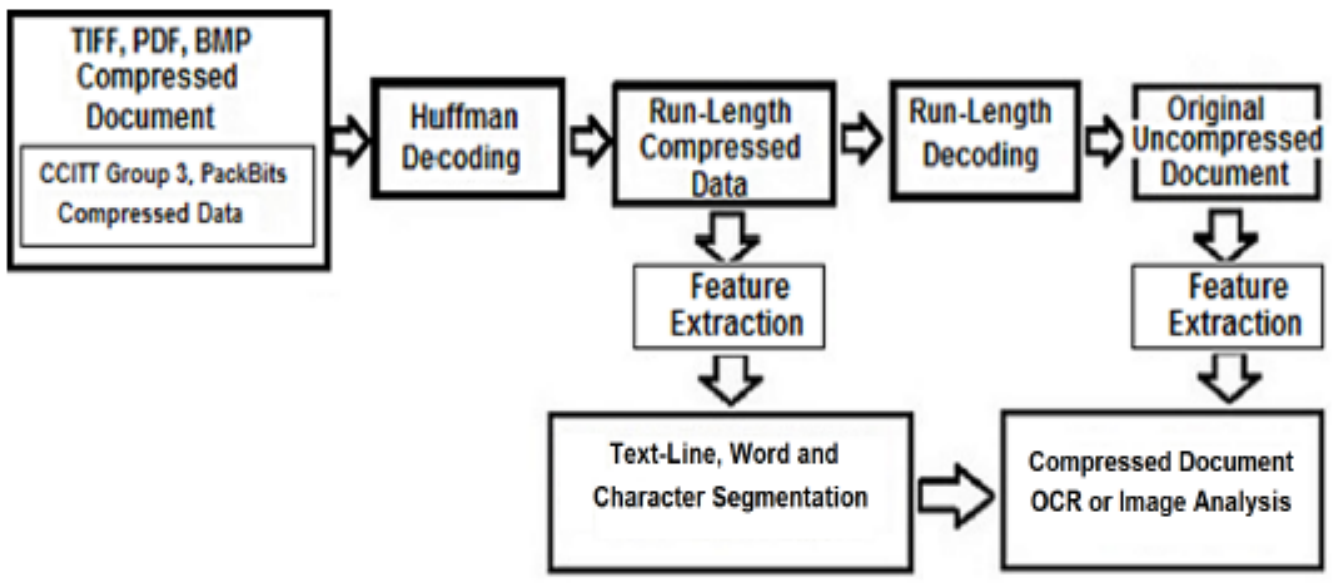

**Fig. 1: Proposed architecture for developing an OCR or Image Analysis using compressed documents**

Overall, our research work aims at exploring the novel idea of directly carrying out segmentation operation from the runlength compressed printed-text-documents. The outcome of this study has the potential to facilitate operations like OCR, word spotting, document equivalence in compressed domain. Detailed discussion regarding these methods and results are placed in later sections. Rest of the paper is organized as follows: section-2 describes the proposed methods for line, word and character segmentation, section-3 presents the experimental results and section-4 concludes the paper with a brief summary.

## II. PROPOSED METHODOLOGY

We begin this section by giving some background information about run-length compressed data and then brief the proposed methods for carrying out segmentation directly in compressed text-document into lines, words and characters.

### *A. Background*

In this research, as we limit our study to text-documents, we choose to work with CCITT Group 3-1D (line by line) coding popularly known as Modified Huffman (MH) coding which is supported by all fax machines. The basic compression technique used here is Run-Length Encoding (RLE). In RLE, a run is a sequence of pixels having similar value and the number of such pixels is length of the run. The representation of the image pixels in the form of sequence of run values (Pi) and its lengths (Li) is run-length encoding. It can be mathematically represented as (P,Li). However, in monochrome images the coding is done with alternate length information of ‘0’ and ‘1’, which produces a compact binary code. The Table-I, gives the description of compressed data using RLE technique. The compressed data consists of alternate columns of number of runs of 0 and 1 identified as odd columns (1, 3, 5,…) and even columns (2, 4, 6,…) respectively. The algorithms developed in this sections work on this type of compressed data.

### *B. Text-Line Segmentation*

Text line segmentation is the first step towards developing a compressed document OCR or any text processing system. However, in order to develop an efficient line segmentation algorithm, it is necessary to understand the compressed data under consideration. In this study, the run-length compressed file consists of alternate columns of runs of white and black pixels. In absence of any black pixel in a row, the compressed row will contain only one run of white pixels whose size will be equal to the width the original document. This special feature of compressed data can be seen in line numbers 1 and 13 of Table-I, which in turn inspires us to use projection profile technique extracted from compressed data [16] for line segmentation in printed text-documents.

A projection profile is a histogram of the number of black pixel values accumulated along parallel lines taken through the document [17]. With our observation, in order to do line segmentation in run-length compressed document, obtaining Horizontal Projection Profile is sufficient, provided that the document is skew free. For an uncompressed document of ‘m’ rows and ‘n’ columns, the mathematical representation for Vertical Projection Profile (VPP) [18] and Horizontal Projection Profile (HPP) are given below as,

$$VPP(y) = \sum_{1 \le x \le m} f(x,y); HPP(x) = \sum_{1 \le y \le n} f(x,y)$$

TABLE I: DESCRIPTION OF RUN-LENGTH COMPRESSED BINARY DATA

| Line | Binary image Data | 1 | 2 | 3 | 4 | 5 |
|---|---|---|---|---|---|---|
| 1: | 00000000000000 | 14 | 0 | 0 | 0 | 0 |
| 2: | 00110000111110 | 2 | 2 | 4 | 5 | 1 |
| 3: | 01111000111110 | 1 | 4 | 3 | 5 | 1 |
| 4: | 01111000111110 | 1 | 4 | 3 | 5 | 1 |
| 5: | 01111000111110 | 1 | 4 | 3 | 5 | 1 |
| 6: | 00110000000000 | 2 | 2 | 10 | 0 | 0 |
| 7: | 10000000000000 | 0 | 1 | 13 | 0 | 0 |
| 8: | 10000000000000 | 0 | 1 | 13 | 0 | 0 |
| 9: | 00100001111100 | 2 | 1 | 4 | 5 | 2 |
| 10: | 01110001111100 | 1 | 3 | 3 | 5 | 2 |
| 11: | 01111001111100 | 1 | 4 | 2 | 5 | 2 |
| 12: | 01111100000000 | 1 | 5 | 8 | 0 | 0 |
| 13: | 00000000000000 | 14 | 0 | 0 | 0 | 0 |

However, the VPP for run run-length compressed data is computed just by adding alternate rows of black runs, thus reducing the number of additions from (mxn) to mx(nˈ/2), where nˈ<n. Making use of minima points in the VPP curve, we trace the vertical start and end points of the text line. The difference between the end point of previous line and start point of current line is difference between the text line. This information along with VPP curve is useful to identify section or subsection title, paragraphs in case of printed documents. The time complexity for the developed algorithm is O(m), where m is the number of rows in a document.

### *C. Word and Character Segmentation*

After successfully segmenting a compressed printed-textdocument into lines, we now think of carrying out word and character segmentation. This stage is more critical while designing an OCR because its accuracy is dependent on precision of segmented words and characters. For printed documents, the common approach used for word and character segmentation is HPP [1]. However in run-length compressed file, tracing of vertical or column information is not straight forward and hence computation of HPP becomes a complex task. Therefore in this study, we propose a novel method for word and character segmentation.

The run-length compressed data consists of alternate runs of white and black pixels. This special data structure motivates us to pull-out the run values from all the rows simultaneously using first two columns. In presence of zero run values in both the columns, the runs on the right are shifted to two positions leftwards. Thus for every pop operation, the run value is decremented by 1 and if it is from first column then the transition value is 0 otherwise 1. This process is repeated for all the rows and generates a sequence of column transitions from run-length compressed file which may be called virtual decompression. The whole idea implies that an imaginary vertical line or column of transitions of 0’s and 1’s is passing through the compressed data from left to right extracting vertical information. This transition information is also useful for obtaining HPP curve. However for run-length compressed documents, this virtual line with zero transitions hints for the presence of space between adjacent columns.

Using this space information we perform character and word segmentation. Once a space between the characters is detected, the spacecounter starts incrementing until a non-space column is detected. If the space-counter is less than the defined wordspace-threshold, then it is identified as a character and entry is made into the character segmentation matrix. In case the condition is false then a word is detected which is stored in word segmentation matrix. This process is repeated until all the columns in a segmented text-line gets exhausted. Here the word-space-threshold has to be computed for every category of documents considered for segmentation. However, in this study we have taken only project-report documents for experimentation. Considering a compressed document of size mxn' where n'< n, the time complexity for the developed algorithm is O(mxn) where n is the width of the original decompressed document. The experimental setup and validation of this algorithm is discussed in Section-3.

The overview of proposed model for line-word-character segmentation of run-length compressed printed English textdocuments is given in Fig-2.

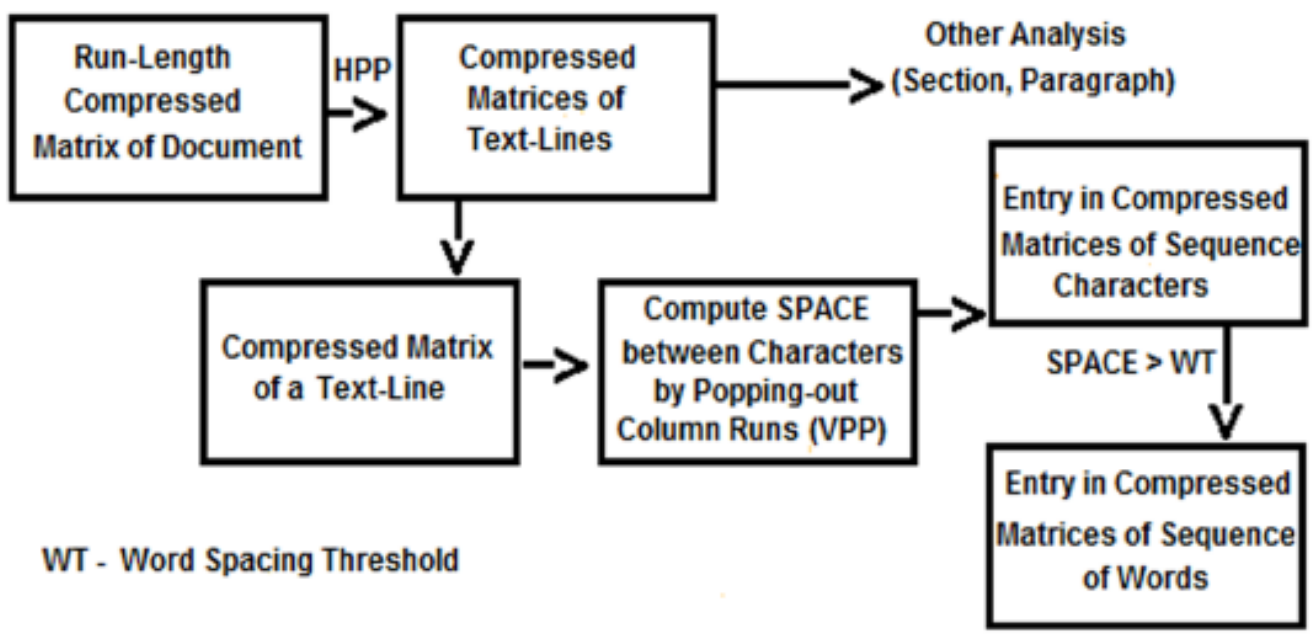

**Fig. 2: Model for the proposed Line-Word-Character segmentation algorithm**

## III. EXPERIMENTAL RESULTS AND DISCUSSIONS

In the previous section, we have proposed algorithms for text line, word and character segmentation from run-length compressed version of documents. The developed algorithms work on run-length compressed data similar to that of in Table-I. In reality, the run-length data has to be extracted from the compressed files using partial huffman decoding shown by Fig-1. In order to validate our proposed algorithms, we could not find any standard data-set of run-length compressed text-documents in the literature. However in this paper, in order to demonstrate and validate our proposed ideas, we have considered 35 single column, noise and skew free, scanned machine printed run-length compressed text-documents (20 Bengali-Scripts, 5 Kannada-Scripts and 10 English-Scripts from project report and journal papers) for experimentation.

A sample document and its segmented text lines in the compressed file are respectively shown in Fig-3a and Fig-3b. The data-set described in the above paragraph contains 1101 text lines. The Precision, Recall and F-measure for text-line segmentation using our proposed method is summarized in Table-II. The precision in case of Bengali script and English script is curtailed respectively due to presence of 21 non-text lines ( page numbers and section-end marker ) and 4 extra segmented lines for mathematical equations.

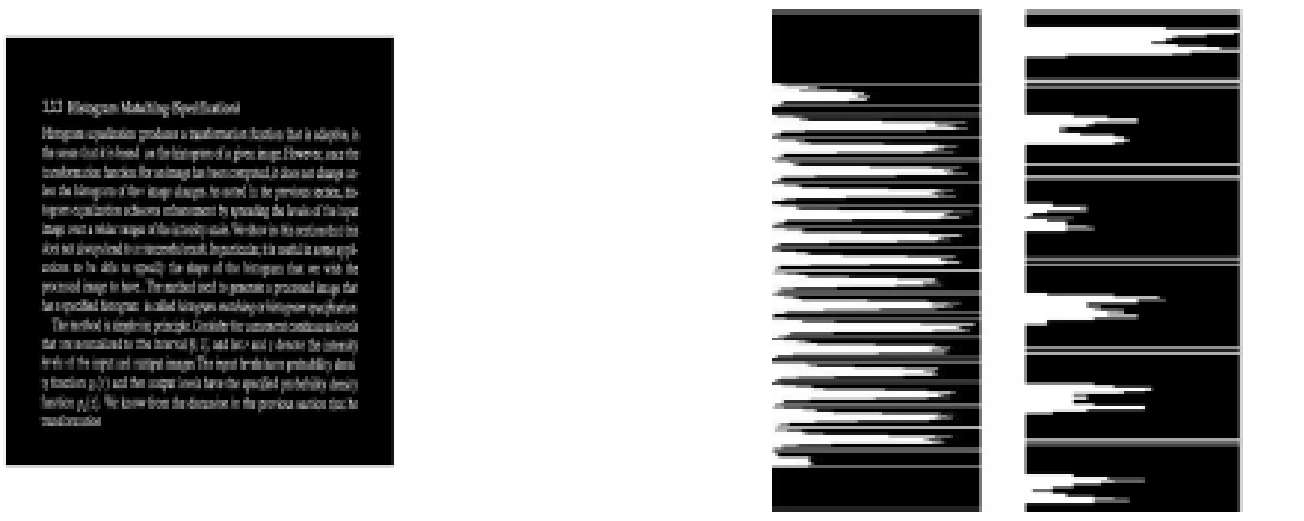

(a) A sample document (b) Zoomed segmented lines

**Fig. 3: Compressed document text-line segmentation**

TABLE II: PRECISION, RECALL AND F-MEASURE USING TEXT-LINE SEGMENTATION ALGORITHM

| Document | No.of Lines | Precision(%) | Recall(%) | F-Measure |
|---|---|---|---|---|
| Bengali Script | 530 | 96.19 | 100 | 98.06 |
| English Script | 434 | 99.09 | 100 | 99.54 |
| Kannada Script | 137 | 100 | 100 | 100 |

The proposed method for word and character segmentation has been designed and developed for English text-documents. However, our word segmentation algorithm can be used to segment words from Bengali and Kannada script documents. Also, the character segmentation algorithm can be applied directly to segment Kannada characters; however for Bengali script it needs slight modifications due to presence of shirorekha in the words. In our study, we consider only 5 English documents of report category for experimentation. The Table-IV shows the extracted compressed positions of sequence of compressed characters for the sample text-line shown in Fig-4. Here, the characters within the columns of bold readings represent the sequence of compressed words. The compressed matrices extracted for characters C6, C7 and C8, which are T, H and E respectively of the word THE shown in Fig-4, are tabulated in Table-V.

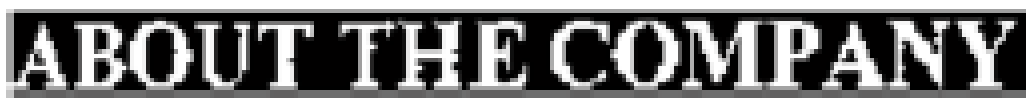

**Fig. 4: A text-line sample**

TABLE III: PRECISION, RECALL AND F-MEASURE FOR 5 ENGLISH SCRIPT DOCUMENTS

| Document | No.of Samples | Precision(%) | Recall(%) | F-Measure |
|---|---|---|---|---|
| Words | 1409 | 96.96 | 99.43 | 98.18 |
| Characters | 7582 | 94.39 | 88.68 | 91.45 |

We also tabulate the Precision, Recall and F-measure obtained for our experiments in Table-III. In this context, the precision for word segmentation is mitigated due to the presence of bullets, numbering and hyphen symbol separated by space on either side which are falsely recognized as words. On the other hand, the performance of character segmentation is reduced drastically because of binarization process just before the document undergoes run-length compression. During this stage, many characters touch and overlap their neighbors of which some of the cases are shown in Fig-5.

TABLE IV: SEGMENTED COMPRESSED MATRICES OF CHARACTERS AND WORDS

| C1 | | C2 | | C3 | | C4 | | C5 | | C6 | | C7 | | C8 | | C9 | | C10 | | C11 | | C12 | | C13 | | C14 | | C15 | |
|---|---|---|---|---|---|---|---|---|---|---|---|---|---|---|---|---|---|---|---|---|---|---|---|---|---|---|---|---|---|
| **1** | 1 | 1 | 1 | 1 | 1 | 1 | 1 | 1 | **1** | **1** | 1 | 1 | 1 | 1 | **1** | **1** | 1 | 1 | 1 | 1 | 1 | 1 | 1 | 1 | 1 | 1 | 1 | 1 | **1** |
| **1** | 1 | 1 | 1 | 1 | 1 | 1 | 1 | 1 | **1** | **1** | 1 | 1 | 1 | 1 | **1** | **1** | 1 | 1 | 1 | 1 | 1 | 1 | 1 | 1 | 1 | 1 | 1 | 1 | **1** |
| **1** | 1 | 1 | 1 | 1 | 3 | 3 | 3 | 3 | **3** | **3** | 5 | 6 | 11 | 11 | **15** | **15** | 17 | 17 | 19 | 20 | 23 | 24 | 25 | 25 | 25 | 26 | 29 | 30 | **33** |
| **1** | 3 | 4 | 5 | 5 | 7 | 8 | 11 | 12 | **13** | **14** | 15 | 15 | 19 | 19 | **21** | **21** | 23 | 23 | 27 | 27 | 31 | 31 | 33 | 33 | 35 | 35 | 39 | 39 | **43** |
| **1** | 3 | 3 | 7 | 7 | 11 | 11 | 15 | 16 | **21** | **22** | 27 | 27 | 31 | 31 | **35** | **35** | 39 | 39 | 43 | 43 | 47 | 47 | 51 | 51 | 53 | 53 | 57 | 57 | **61** |
| **1** | 3 | 3 | 7 | 7 | 11 | 11 | 15 | 16 | **21** | **22** | 27 | 27 | 31 | 31 | **35** | **35** | 39 | 39 | 43 | 43 | 49 | 49 | 53 | 53 | 55 | 55 | 59 | 59 | **63** |
| **1** | 3 | 3 | 7 | 7 | 11 | 11 | 15 | 16 | **19** | **19** | 21 | 21 | 25 | 25 | **31** | **31** | 35 | 35 | 39 | 39 | 47 | 47 | 51 | 51 | 53 | 53 | 57 | 57 | **61** |
| **1** | 3 | 3 | 7 | 8 | 11 | 11 | 15 | 15 | **17** | **17** | 19 | 19 | 23 | 23 | **27** | **28** | 31 | 31 | 35 | 35 | 43 | 43 | 47 | 47 | 49 | 49 | 55 | 55 | **59** |
| **1** | 3 | 3 | 7 | 8 | 11 | 11 | 15 | 15 | **17** | **17** | 19 | 19 | 23 | 23 | **27** | **28** | 29 | 29 | 33 | 33 | 41 | 41 | 45 | 45 | 47 | 47 | 53 | 53 | **57** |
| **1** | 5 | 5 | 9 | 10 | 13 | 13 | 17 | 17 | **19** | **19** | 21 | 21 | 23 | 23 | **25** | **26** | 27 | 28 | 31 | 31 | 39 | 39 | 43 | 43 | 47 | 47 | 53 | 53 | **55** |
| **1** | 5 | 5 | 7 | 8 | 11 | 11 | 15 | 15 | **17** | **17** | 19 | 19 | 21 | 21 | **23** | **24** | 25 | 26 | 29 | 29 | 37 | 37 | 39 | 39 | 43 | 43 | 47 | 47 | **49** |
| **1** | 5 | 5 | 9 | 10 | 13 | 13 | 17 | 17 | **19** | **19** | 21 | 21 | 25 | 25 | **29** | **30** | 31 | 31 | 35 | 35 | 41 | 41 | 43 | 43 | 47 | 47 | 51 | 51 | **53** |
| **1** | 5 | 5 | 9 | 10 | 13 | 13 | 17 | 17 | **19** | **19** | 21 | 21 | 25 | 25 | **29** | **30** | 31 | 31 | 35 | 35 | 41 | 41 | 43 | 43 | 45 | 45 | 49 | 49 | **51** |
| **1** | 3 | 3 | 7 | 8 | 11 | 11 | 15 | 15 | **17** | **17** | 19 | 19 | 23 | 23 | **29** | **30** | 31 | 31 | 35 | 35 | 41 | 41 | 43 | 43 | 45 | 45 | 49 | 49 | **51** |
| **1** | 5 | 5 | 9 | 9 | 13 | 13 | 17 | 17 | **19** | **19** | 21 | 21 | 25 | 25 | **29** | **29** | 33 | 33 | 37 | 37 | 43 | 43 | 45 | 45 | 49 | 49 | 53 | 53 | **55** |
| **1** | 5 | 5 | 9 | 9 | 13 | 13 | 17 | 17 | **19** | **19** | 21 | 21 | 25 | 25 | **29** | **29** | 33 | 33 | 37 | 37 | 43 | 43 | 45 | 45 | 49 | 49 | 53 | 53 | **55** |
| **1** | 5 | 5 | 9 | 9 | 13 | 13 | 17 | 17 | **19** | **19** | 21 | 21 | 25 | 25 | **29** | **29** | 33 | 33 | 37 | 37 | 43 | 43 | 45 | 45 | 49 | 49 | 53 | 53 | **55** |
| **2** | 5 | 6 | 7 | 7 | 9 | 9 | 11 | 11 | **13** | **13** | 15 | 16 | 19 | 20 | **21** | **21** | 23 | 23 | 25 | 26 | 29 | 30 | 31 | 32 | 35 | 36 | 39 | 39 | **41** |
| **1** | 1 | 1 | 1 | 1 | 1 | 1 | 1 | 1 | **1** | **1** | 1 | 1 | 1 | 1 | **1** | **1** | 1 | 1 | 1 | 1 | 1 | 1 | 1 | 1 | 1 | 1 | 1 | 1 | **1** |

TABLE V: RUN-LENGTH COMPRESSED MATRICES OF CHARACTERS OF THEWORD *THE*

| Run-Length Compressed Data | | | | | | | | | | | | | | | | | | | | |
|---|---|---|---|---|---|---|---|---|---|---|---|---|---|---|---|---|---|---|---|---|
| **C6 (T)** | | | | | | | **C7 (H)** | | | | | | | **C8 (E)** | | | | | | |
| 50 | 0 | 0 | 0 | 0 | 0 | 0 | 50 | 0 | 0 | 0 | 0 | 0 | 0 | 50 | 0 | 0 | 0 | 0 | 0 | 0 |
| 50 | 0 | 0 | 0 | 0 | 0 | 0 | 50 | 0 | 0 | 0 | 0 | 0 | 0 | 50 | 0 | 0 | 0 | 0 | 0 | 0 |
| 12 | 1 | 2 | 0 | 0 | 0 | 0 | 0 | 1 | 8 | 1 | 4 | 2 | 4 | 2 | 1 | 5 | 4 | 15 | 0 | 0 |
| 0 | 14 | 2 | 0 | 0 | 0 | 0 | 1 | 5 | 4 | 5 | 6 | 0 | 0 | 3 | 11 | 11 | 0 | 0 | 0 | 0 |
| 0 | 2 | 3 | 4 | 3 | 1 | 4 | 2 | 3 | 6 | 4 | 6 | 0 | 0 | 3 | 4 | 5 | 2 | 9 | 0 | 0 |
| 0 | 1 | 4 | 3 | 4 | 1 | 4 | 2 | 3 | 6 | 4 | 6 | 0 | 0 | 3 | 4 | 6 | 1 | 8 | 0 | 0 |
| 5 | 3 | 9 | 0 | 0 | 0 | 0 | 2 | 3 | 6 | 4 | 6 | 0 | 0 | 3 | 4 | 3 | 1 | 2 | 1 | 8 |
| 5 | 3 | 9 | 0 | 0 | 0 | 0 | 2 | 3 | 6 | 4 | 6 | 0 | 0 | 3 | 4 | 3 | 1 | 10 | 0 | 0 |
| 5 | 4 | 8 | 0 | 0 | 0 | 0 | 2 | 3 | 6 | 4 | 6 | 0 | 0 | 3 | 4 | 2 | 2 | 10 | 0 | 0 |
| 5 | 3 | 9 | 0 | 0 | 0 | 0 | 2 | 12 | 7 | 0 | 0 | 0 | 0 | 3 | 8 | 10 | 0 | 0 | 0 | 0 |
| 5 | 3 | 9 | 0 | 0 | 0 | 0 | 2 | 13 | 6 | 0 | 0 | 0 | 0 | 3 | 8 | 10 | 0 | 0 | 0 | 0 |
| 5 | 3 | 9 | 0 | 0 | 0 | 0 | 2 | 3 | 6 | 4 | 6 | 0 | 0 | 3 | 4 | 2 | 2 | 10 | 0 | 0 |
| 5 | 3 | 9 | 0 | 0 | 0 | 0 | 2 | 3 | 6 | 3 | 7 | 0 | 0 | 3 | 4 | 3 | 1 | 10 | 0 | 0 |
| 5 | 3 | 9 | 0 | 0 | 0 | 0 | 2 | 4 | 5 | 3 | 7 | 0 | 0 | 3 | 4 | 3 | 1 | 3 | 1 | 6 |
| 5 | 3 | 9 | 0 | 0 | 0 | 0 | 2 | 3 | 6 | 4 | 6 | 0 | 0 | 3 | 4 | 7 | 1 | 7 | 0 | 0 |
| 5 | 3 | 9 | 0 | 0 | 0 | 0 | 2 | 3 | 6 | 4 | 6 | 0 | 0 | 3 | 4 | 6 | 2 | 8 | 0 | 0 |
| 4 | 5 | 7 | 0 | 0 | 0 | 0 | 1 | 5 | 4 | 5 | 6 | 0 | 0 | 3 | 5 | 3 | 4 | 9 | 0 | 0 |
| 3 | 7 | 5 | 0 | 0 | 0 | 0 | 0 | 7 | 2 | 7 | 2 | 0 | 0 | 14 | 12 | 0 | 0 | 0 | 0 | 0 |
| 50 | 0 | 0 | 0 | 0 | 0 | 0 | 50 | 0 | 0 | 0 | 0 | 0 | 0 | 50 | 0 | 0 | 0 | 0 | 0 | 0 |

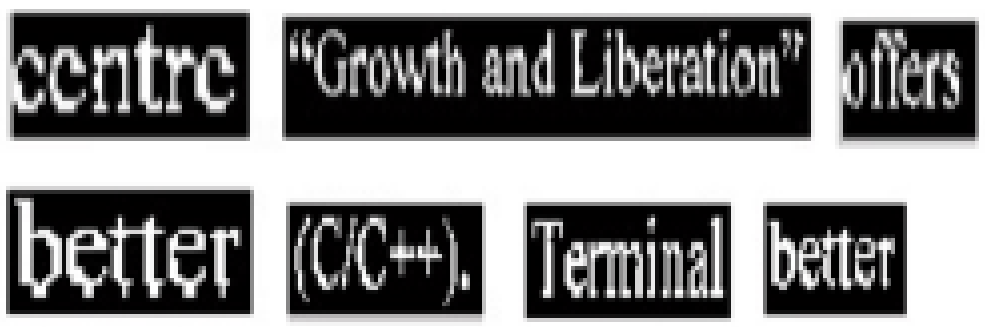

**Fig. 5: Touching and overlapping characters due to binarization**

In all, the experimental results presented in this section validate our proposed methods for line, word and character segmentation using compressed documents. As an outcome of this research work, we are motivated to apply these algorithms further for document understanding, equivalence [19] and pattern matching problems using compressed data.

## IV. CONCLUSION

In this research work, we have proposed two novel algorithms for segmentation of compressed text-documents into lines, words and characters which is considered to be critical pre-processing stage in developing an OCR for compressed documents. The proposed algorithms work on runlength compressed data and have been validated taking 35 compressed documents (20 Bengali-Scripts, 5 Kannada-Scripts and 10 English-Scripts from project report and journal papers) giving satisfactory results. Further, as a result of this research work we are motivated to think of developing applications that facilitate operations like OCR, word spotting, document equivalence using compressed files.

## REFERENCES

[1] N. Priyanka, S. Pal, and R. Mandal, “Line and word segmentation approach for printed documents,” IJCA Special Issue on Recent Trends in Image Processing and Pattern Recognition”, pp. 30–36, 2010.

[2] N. Nikolaou, M. Makridis, B. Gatos, N. Stamatopoulos, and N. Papamarkos, “Segmentation of historical machine-printed documents using adaptive run length smoothing and skeleton segmentation paths,” Image and Vision Computing, vol. 28, pp. 590–604, 2010.

[3] K. R. McConnell, D. Bodson, and R. Schaphorst, FAX: Digital Facsimile Technology and Applications. Artech House, 1989.

[4] G. Grant and A. Reid, “An efficient algorithm for boundary tracing and feature extraction,” Computer Graphics and Image Processing, vol. 17, pp. 225–237, November 1981.

[5] T. Tsuiki, T. Aoki, and S. Kino, “Image processing based on a runlength coding and its application to an intelligent facsimile,” Proc. Conf.Record, GLOBECOM ’82, pp. B6.5.1–B6.5.7, November 1982.

[6] J. Capon, “A probabilistic model for run-length coding of pictures,” IRE Transactions on Information Theory, vol. 5, pp. 157–163, 1959

[7] .J. Limb and I. Sutherland, “Run-length coding of television signals,” Proceedings of IEEE, vol. 53, pp. 169–170, 1965.

[8] Y. Shima, S. Kashioka, and J. Higashino, “A high-speed algorithm for propagation-type labeling based on block sorting of runs in binary images,” Proceedings of 10th International Conference on Pattern Recognition (ICPR), vol. 1, pp. 655–658, 1990.

[9] E. Regentova, S. Latifi, S. Deng, and D. Yao, “An algorithm with reduced operations for connected components detection in itu-t group 3/4 coded images,” IEEE Transactions on Pattern Analysis and Machine Intelligence, vol. 24, pp. 1039 – 1047, August 2002.

[10] A. L. Spitz, “Analysis of compressed document images for dominant skew, multiple skew, and logotype detection,” Computer vision and Image Understanding, vol. 70, pp. 321–334, June 1998.

[11] E. Regentova, S. Latifi, D. Chen, K. Taghva, and D. Yao, “Document analysis by processing jbig-encoded images,” International Journal on Document Analysis and Recognition (IJDAR), vol. 7, pp. 260–272, 2005.

[12] D. S. Lee and J. J. Hull, “Detecting duplicates among symbolically compressed images in a large document database,” Pattern Recognition Letters, vol. 22, pp. 545–550, 2001.

[13] J. J. Hull, “Document image similarity and equivalence det,” International Journal on Document Analysis and Recognition (IJDAR’98), vol. 1, pp. 37–42, 1998.

[14] Y. Lu and C. L. Tan, “Document retrieval from compressed images,” Pattern Recognition, vol. 36, pp. 987–996, 2003.

[15] T. M. Breuel, “Binary morphology and related operations on run-length representations,” International Conference on Computer Vision Theory and Applications - VISAPP, pp. 159–166, 2008.

[16] M. Javed, P. Nagabhushan, and B. B. Chaudhuri, “Extraction of projection profile, run-histogram and entropy features straight from run-length compressed documents,” Proceedings of Second IAPR Asian Conference on Pattern Recognition (ACPR’13), Okinawa, Japan, November 5-8, 2013 in Press.

[17] R. Kasturi, L. O. Gorman, and V. Govindaraju, “Document imageanalysis: A primer,” Sadhana Part 1, vol. 27, pp. 3–22, 2002.

[18] L. Likforman-Sulem, A. Zahour, and B. Taconet, “Text line segmentation of historical documents: a survey,” International Journal of Document Analysis and Recognition (IJDAR), vol. 9, pp. 123–138, April 2007.

[19] S. D. Gowda and P. Nagabhushan, “Entropy quantifiers useful for establishing equivalence between text document images,” International Conference on Computational Intelligence and Multimedia Applications, pp. 420 – 425, 2007.